\documentclass{article}

\usepackage{arxiv}

\usepackage[utf8]{inputenc} 
\usepackage[T1]{fontenc}    
\usepackage{hyperref}       
\usepackage{url}            
\usepackage{booktabs}       
\usepackage{amsfonts}       
\usepackage{nicefrac}       
\usepackage{microtype}      
\usepackage{lipsum}
\usepackage{graphicx}
\usepackage{caption}
\usepackage{float}
\usepackage{dsfont}
\usepackage{graphicx}
\usepackage{xcolor}
\usepackage{mathtools}
\usepackage{amsmath}

\usepackage{amssymb}
\usepackage{xcolor}
\usepackage{authblk}
\usepackage{colortbl}
\usepackage{multicol}
\usepackage{multirow}
\usepackage[backend=biber,style=ieee]{biblatex}
\bibliography{references}
\usepackage{fontawesome}
\usepackage{hyperref}
\usepackage{fancyhdr}
\usepackage{latexsym}
\usepackage{framed,multirow}
\usepackage{subcaption}  
\usepackage[bottom]{footmisc}
\newcommand\blfootnote[1]{%
  \begingroup
  \renewcommand\thefootnote{}\footnote{#1}%
  \addtocounter{footnote}{-1}%
  \endgroup
}
\newcommand*\samethanks[1][\value{footnote}]{\footnotemark[Equal contribution]}

\usepackage{latexsym}

\title{In-Context Ensemble Learning from Pseudo Labels Improves Video-Language Models for Low-Level Workflow Understanding}

\author[1]{Moucheng Xu* }
\author[1]{Evangelos Chatzaroulas*}
\author[1]{Luc McCutcheon}
\author[1]{Abdul Ahad} 
\author[1]{Hamzah Azeem}
\author[1]{Janusz Marecki}
\author[1]{Ammar Anwar}

\affil[1]{Agile Loop, London, UK}
\affil[*]{Equal contribution}
\affil[\faPhone]{Contact: moucheng.xu.18@alumni.ucl.ac.uk}
\affil[\faDesktop]{\href{https://github.com/moucheng2017/action-labelling}{Code URL link}}

\begin{document}

\maketitle

\begin{abstract}
A Standard Operating Procedure (SOP) defines a low-level, step-by-step written guide for a business software workflow. SOP generation is a crucial step towards automating end-to-end software workflows. Manually creating SOPs can be time-consuming. Recent advancements in large video-language models offer the potential for automating SOP generation by analyzing recordings of human demonstrations. However, current large video-language models face challenges with zero-shot SOP generation. In this work, we first explore in-context learning with video-language models for SOP generation. We then propose an exploration-focused strategy called In-Context Ensemble Learning, to aggregate pseudo labels of multiple possible paths of SOPs. The proposed in-context ensemble learning as well enables the models to learn beyond its context window limit with an implicit consistency regularisation. We report that in-context learning helps video-language models to generate more temporally accurate SOP, and the proposed in-context ensemble learning can consistently enhance the capabilities of the video-language models in SOP generation.
\end{abstract}

\keywords{Multimodal In-Context Learning \and Large Video-Language Models \and Ensemble Learning \and SOP generation \and Pseudo Labels \and Test Time Tree Ensemble}

\section{Introduction}
Video-language models are an emerging family of large foundational models attracting growing research interest. These models typically pre-train a visual encoder to project visual inputs into tokens, which are then utilised by large language models to interpret visual signals alongside text instructions. Recent video-language models have achieved significant success in high-level video understanding, such as high-level video summarisation, as reflected in their strong performance across diverse VQA benchmarks \cite{maaz-etal-2024-video,Lin_2024_CVPR,Chen_2024_CVPR}. However, current video-language models still face challenges when it comes to more complicated tasks. For instance, most available models cannot handle long videos or multiple short videos. Another challenge is their lack of competence in complex low-level video understanding\cite{wornow2024multimodal}. Standard Operating Procedure (SOP) generation is a typical low-level video understanding task \cite{wornow2024multimodal} from videos. SOP involves detailed, step-by-step descriptions of the chronological ordering of actions taken in a software workflow on a recorded screen. The automatic generation of accurate SOPs can pave the way for the automation of business software. \blfootnote{To appear in NeurIPS Workshop on Video-Language Models 2024} 

In this work, we focus on improving the frontier video-language models in SOP generation from two perspectives: fine-tuning and increasing test-time compute. Since a lot of business software are domain specific, SOP generation is a novel task to the large models. One promising paradigm to fine-tune pre-trained large models to new unseen tasks is to use In-Context Learning \cite{NEURIPS2020_1457c0d6,pmlr-v202-von-oswald23a,pan-etal-2023-context,akyrek2023what,NEURIPS2023_0561738a,wei2022emergent,wei2022finetuned}. In-Context Learning includes both the training samples and testing questions during the prompt. Unlike supervised fine-tuning, in-context learning requires no model parameter updates. In our case, we need to include training videos as a part of the multimodal prompt. However, in our preliminary exploratory experiments, we found that video-language models cannot receive as many video-training samples as the text- or images-training counterparts. This leads to an issue that the models hit the context window limit before it sees enough number of training videos, hampering the usability of in-context learning. Apart from fine-tuning, we also focus on extending the test-time computation. This is a common strategy for complicated reasoning tasks \cite{wang2023selfconsistency,wei2022chain,yao2023tree,NEURIPS2023_91edff07,snell2024scaling,khan24a} based on the intuition that humans normally spend more time generating multiple plausible reasoning paths before deciding the next action in complicated cognitive tasks. In order to increase the test-time compute without violating the context window limit at the same time, we present a new in-context learning strategy called In-Context Ensemble (ICE) Learning to achieve both the targets simultaneously. In summary, our main contributions of this paper can be summarized into two-fold: 
\begin{enumerate}
    \item We present the first evaluation of multimodal in-context learning in SOP generation with video-language models. We report that in-context learning consistently helps models to predict the step-by-step actions in a more accurate temporal order.
    \item We introduce a multimodal in-context ensemble learning approach to increase the test-time compute of the video-language models in SOP generation with pseudo-labels. The in-context ensemble learning enables the models to learn from more video examples beyond their context window limit, with a self regularisation effect. We report that the proposed multimodal in-context ensemble learning consistently improves the models' predictions. 
\end{enumerate}

\section{In-Context Ensemble (ICE) Learning}
\label{sec:ice}
\begin{figure}[H]
\centering
\includegraphics[width=\textwidth]{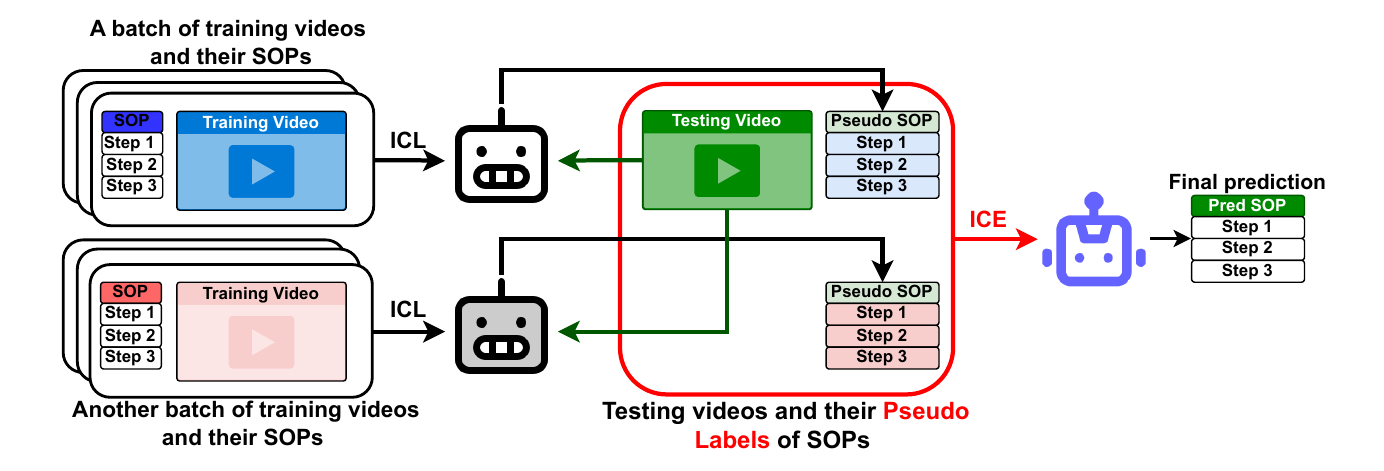}
\caption{The proposed Multimodal In-Context Ensemble (ICE) learning with pseudo-labels. ICL: in-context learning. SOP: standard operating procedure, detailing chronological step-by-step actions in the video.}
\label{fig:ice}
\end{figure}
By combining pseudo labelling \cite{Lee2013PseudoLabelT,XU2024103125,10.1007/978-3-031-16443-9_56,XU2024103125,asano2020self} with In-Context Learning (ICL) \cite{NEURIPS2020_1457c0d6,pmlr-v202-von-oswald23a,pan-etal-2023-context,akyrek2023what,NEURIPS2023_0561738a,wei2022emergent,wei2022finetuned}, we present Multimodal In-Context Ensemble (ICE) learning to enhance video-language models' understanding of low-level actions in videos. An illustration of the ICE pipeline is shown in above Fig \ref{fig:ice}. Our ICE pipeline first applies multimodal ICL to train several video-language models on different batches of training videos, along with their gold standard SOP text labels. Those fine-tuned video-language models then create different pseudo labels of SOPs from the testing videos. Finally, another video-language model processes the testing videos, along with those different pseudo labels as priors, to generate the final predictions of the SOPs. The proposed ICE pipeline offers four key benefits:

\begin{enumerate}

    \item The introduction of the pseudo labelling in the reasoning process linearly scales up the test-time compute of the video-language models, improving the models' reasoning ability. Our ICE can be seen as an analogy to a black-box version of test-time tree search for aggregating multiple potential reasoning paths. But our ICE is much simpler than the test-time tree search methods.
        
    \item Secondly, the ICE stage enables the model to learn from more examples beyond its context window limit. For instance, during our early exploration, we could only feed up to 8 high-resolution videos as training examples to GPT-4o-mini due to the context window length. However, with ICE, in our experiments, our GPT-4o-mini can successfully see the equivalent of 24 videos as training examples.

    \item The third benefit is that the provided pseudo-labels sometimes share similar actions with variations. This implicit self consistency regularisation also aids the model's reasoning. 
    
    \item Lastly, the model does not need to utilise the entire context window to store the training video examples in the prompt, allowing it to focus more on the testing video. This is to address a common drawback of the large models that they struggles with information retrieval in extremely long tokens (e.g. needle search in a haystack) \cite{kuratov2024search,NEURIPS2020_6b493230,gu2024mambalineartimesequencemodeling,ye2024differential}.
    
\end{enumerate}

\section{Related Work}

\textbf{Ensemble} Ensemble scales the inference compute by treating each candidate prediction equally to improve the base model's reasoning to reach a more accurate final prediction. The most common strategy of ensemble is majority voting. Majority voting picks up the most frequent prediction of all of the candidate predictions as the final prediction \cite{wang2023selfconsistency,chen2024alphamath,snell2024scaling}. A lot of existing large models (e.g. Google Gemini, OpenAI GPT4) provide options for enabling majority voting during inference. We also use the majority voting ensemble as a baseline in our experiments. Different from traditional majority voting ensemble, our proposed ICE has a proposal sampling distribution with large variances. This is because the traditional majority voting ensemble samples candidate predictions from the same model \cite{snell2024scaling}, whereas our ICE samples per candidate prediction from each differently fine-tuned model. 

\textbf{Consistency Prompts} Consistency regularisation has been playing a pivotal role in semi-supervised and self-supervised learning \cite{Chen_2021_CVPR,midl2022_xu,fixmatch2020,tmi2023_xu}. The general idea is to apply a consistency restraint on the predictions from different experts that are fed with the same source input. A recent work \cite{roy2024consistencyguided} also discovered that consistency regularisation improves the video-language models at few-shot learning while the inputs prompts are differently perturbed. Although our In-Context Ensemble (ICE) does not specifically implement the consistency regularisation. By learning from different pseudo labels, our ICE implicitly enforces consistency regularisation, because sometimes, the same actions do appear among different pseudo labels but with variations.

\textbf{Test-Time Tree Search} Tree search has been a popular tool to scale up test-time compute by aggregating the candidate sequential reasoning processes \cite{chen2024alphamath,snell2024scaling,feng2023alphazerolike,DBLP:journals/corr/abs-1712-01815} for better complex reasoning. From the end-to-end perspective, our ICE shares similiarities with Test-Time Tree Search, especially Lookahead Search. Each SOP can be seen as a sequential reasoning process, changing the status of the screen from the first frame to the last frame, following a particular temporal path. Multiple SOP pseudo labels resemble a tree structure and the generation of the final SOP is a black-box version of steps aggregation of pseudo labels, thereby an analogy to Tree Search methods. However, there are noticeable differences between the Tree Search methods and our ICE. In Beam Search or Lookahead Search, a Verifier \cite{lightman2024lets,wang-etal-2024-math,DBLP:journals/corr/abs-2110-14168} is used to select the action step-wise, at each level of the tree, focusing on exploiting the options. Whereas in our ICE, we use a large video-language model to consider the pseudo labels to generate the final chain-of-the-steps, focusing on conditioning exploration, which is beyond exploiting available candidate pseudo labels. In addition, we do not train an extra Verifier and we do not have an explicit selection process per action.  

\textbf{Reflection Prompts} Self-revision has also been another popular tool to scale up test-time compute by iteratively prompting the same large language model to refine its initial prediction \cite{NEURIPS2023_91edff07,saunders2022selfcritiquing,khan24a}. This approach is simple yet effective, computationally affordable too. In our ICE method, the models are implicitly asked to do self-reflection on the old predictions, which are the pseudo labels. However, different from the single model based reflection \cite{NEURIPS2023_91edff07,saunders2022selfcritiquing}, our ICE involved with multiple models which are respectively fine-tuned on different data. We argue that by doing so, we can increase the variance of the proposal distribution for better reasoning outcomes. 

\section{Experiments}
\textbf{Data} The WONDERBREAD benchmark \cite{wornow2024multimodal} is a recently curated benchmark containing 2928 videos of distinct workflows and the detailed SOPs of the chronological step-by-step actions in the videos. We use a subset called "Gold Demo" from the WONDERBREAD benchmark of 724 videos. We randomly split the data into two parts, 30\% training (217 videos) and 70\% testing (507 videos). We subsequently split the training data into 28 batches, each batch contains 8 videos. Due to the limited computational resources, we only used the first 3 training batches. 

\textbf{Baseline 1: zero-shot} We aim to evaluate the effectiveness of in-context learning with video-language models in SOP generation. The original zero-shot evaluation in the WONDERBREAD benchmark \cite{wornow2024multimodal} injects additional information such as intent (a high-level summary of the video) and action trace (a detailed log of all user interactions with the screen, including elements' coordinates). We argue that providing such additional information to the models may not necessarily reflect the actual reasoning abilities of the foundational models. Therefore, we only feed the models the raw visual information, namely the key frames of the videos. 

\textbf{Baseline 2: ensemble} We want to compare our multimodal in-context ensemble against traditional ensemble with video-language models in SOP generation. Therefore, our second baseline is a majority voting ensemble of the pseudo labels in text using the models, referred to as "Ensemble".  We prompt the large models (GPT-4o-mini and Gemini-1.5-flash) to do automatic majority voting ensemble of the pseudo labels. 

\textbf{Video-Language Models} As our approach requires long context windows with supports for multiple rounds of conversations, our choices of the models are limited to mostly closed models. Further considering the trade-off between performance and cost, we use GPT-4o-mini (OpenAI) and Gemini-1.5-flash (Google). We also include a zero-shot evaluation of a very recent public model from Microsoft, Phi-3.5. Throughout the experiments, to accommodate the context length window limits, if a video is longer than 20 frames, we down-sample the video to 20 frames. We also keep the videos at a high resolution of 1024 x 768.

\textbf{Metrics} Precision measures how many steps in the prediction match those in the gold standard SOP. Recall measures how many steps of the gold standard SOP are included in the predictions. Temporal order evaluates whether the sequence of steps in the prediction aligns with the sequence in the gold standard SOP. All evaluations are performed using GPT-4o-mini, following the original guidelines in \cite{wornow2024multimodal}.

\section{Results}
\begin{table}[H]
\centering
\begin{tabular}{|c|c|c|c|c|}
\hline
\textbf{Method} & \textbf{Training Data} & \textbf{Precision (\%) $\uparrow$} & \textbf{Recall (\%) $\uparrow$} & \textbf{Temporal Order (\%) $\uparrow$} \\ 
\hline
\multicolumn{5}{|c|}{\textbf{GPT-4o mini}} \\ 
\hline
zero-shot & n/a & 42.62 & 78.13 & 32.93 \\ 
8-shot ICL & Batch 1 & 43.16 & 78.13 & 33.46 \\ 
8-shot ICL & Batch 2 & \textbf{\textcolor{red}{45.95 (+3.33)}} & 78.13 & 34.15 \\ 
8-shot ICL & Batch 3 & 44.51 & 78.13 & 33.08 \\ 
Ensemble & Batch 1 - 3 & 36.05 & 72.31 & 21.47 \\ 
\rowcolor{gray!20} 
ICE & Batch 1 - 3 & 44.34 (+1.72) & \textbf{\textcolor{red}{84.79 (+6.66)}} & \textbf{\textcolor{red}{37.17 (+4.24)}} \\ 
\hline
\multicolumn{5}{|c|}{\textbf{Gemini-1.5 flash}} \\ 
\hline
zero-shot & n/a & 34.35 & 45.16 & 27.82 \\ 
8-shot ICL & Batch 1 & 34.10 & 45.18 & 35.15 \\ 
8-shot ICL & Batch 2 & 33.99 & 41.96 & 29.42 \\ 
8-shot ICL & Batch 3 & 34.08 & 40.75 & 29.77 \\ 
24-shot ICL & Batch 1 - 3 & 29.75 & 39.42 & 26.47 \\ 
Ensemble & Batch 1 - 3 & 30.30 & 40.52 & 26.12 \\ 
\rowcolor{gray!20} 
ICE & Batch 1 - 3 & \textbf{\textcolor{red}{40.77 (+6.42)}} & \textbf{\textcolor{red}{54.38 (+9.22)}} & \textbf{\textcolor{red}{35.89 (+8.07)}} \\ 
\hline
\multicolumn{5}{|c|}{\textbf{GPT-4o mini + Gemini-1.5 flash}} \\ 
\hline
\rowcolor{gray!20} 
ICE & Batch 1 - 3 & 41.54 & 83.34 & 34.33 \\ 
\hline
\multicolumn{5}{|c|}{\textbf{Phi-3.5}} \\ 
\hline
zero-shot & n/a & 31.66 & 45.88 & 24.42 \\ 
\hline
\end{tabular}
\caption{Testing results on 507 videos from the ``Gold Demo'' subset of WONDERBREAD benchmark\cite{wornow2024multimodal}. Our evaluation is more challenging than the original one in \cite{wornow2024multimodal} as we only feed videos into the models, \textit{without} trace information (e.g. mouse clicks). ICL: in-context learning. ICE: in-context ensemble. Ensemble: majority voting of pseudo labels}
\label{tab:results}
\end{table}

As shown in Tab.~\ref{tab:results}, we observe that GPT-4o-mini generally outperforms Gemini-1.5-flash. And Gemini-1.5-flash slightly surpasses Phi-3.5. In-context learning (ICL) consistently improves the models' ability to predict more correct order of steps compared to zero-shot baselines across different settings. ICE further enhances the models' performance for both GPT4o-mini and Gemini-1.5-flash. Notably, for Gemini-1.5-flash, ICE resulted in a 9.22\% improvement in recall. With GPT-4o-mini, ICE achieves high recall at 84.79\% but low precision at 44.34\%, suggesting that, while the model can successfully predict most of the steps, it may have over-imputed too many intermediate steps that were omitted in the ground truth. ICE also significantly outperforms the majority voting "Ensemble", highlighting the necessity of multimodal training samples in SOP generation. Further investigations are required to understand why "Ensemble" underperforms its individual components. Interestingly, the 24-shot ICL experiment with Gemini-1.5-flash shows that the model's improvements do not appear to increase linearly with the number of training samples. This may suggest that the foundational model naturally struggles to attend to too many high-resolution images at once. We also include a successful case and a failed case in the Appendix \ref{appendix}.

\section{Analysis}
\begin{figure}[H]
\centering
\includegraphics[width=0.9\textwidth]{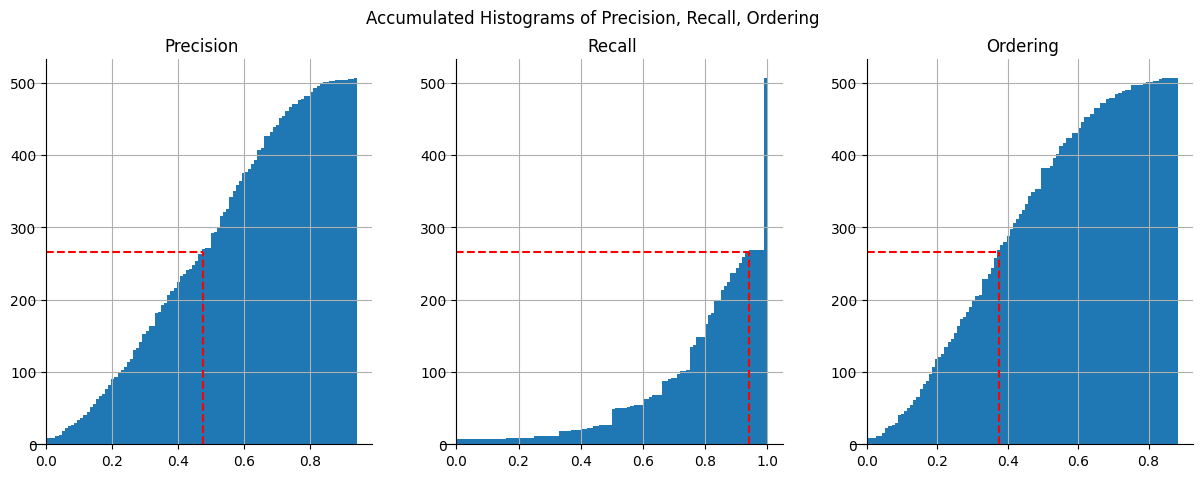}
\caption{Accumulated histograms of testing results with the ICE with GPT-4o-mini. The red vertical lines indicate the 50-percentile of testing cases.}
\label{fig:accumulated_histograms}
\end{figure}
We visualised the accumulated histograms of the metrics in the above Fig.\ref{fig:accumulated_histograms}. The 50th percentile of recall is above 90\%, showing that the model is capable at predicting the actions in the ground truth. However, the 50th percentile of precision is lower than 50\%, and the 50th percentile of temporal ordering is even lower at, less than 40\%. We hypothesise that the unsatisfactory performance in precision and ordering metrics might be due to the model having its own preferences for the style in which the output steps are written. To explore this further, we examined the relationship between the lengths of the SOPs and the model's performance, as SOP length reflects the model’s preferred writing style.

\begin{figure}[H]
    \centering
        \includegraphics[width=0.6\textwidth]{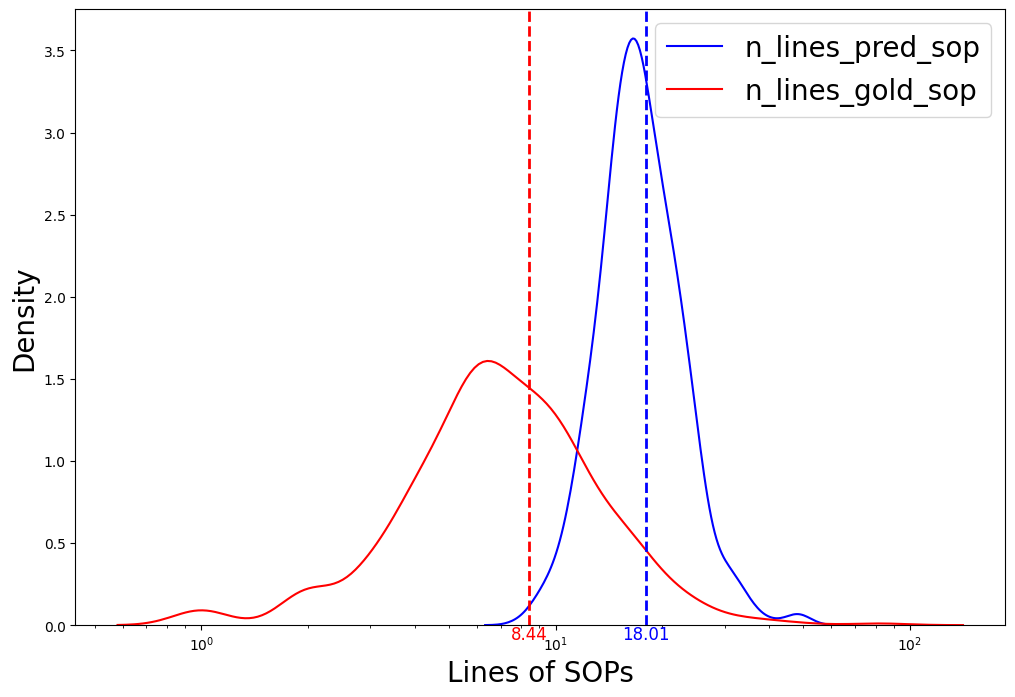}
        \caption{Kernel density estimate plots of lines of SOPs. ICE with GPT-4o-mini.}
        \label{fig:lines_sops}
\end{figure}

\begin{figure}[H]
        \centering
        \includegraphics[width=0.6\textwidth]{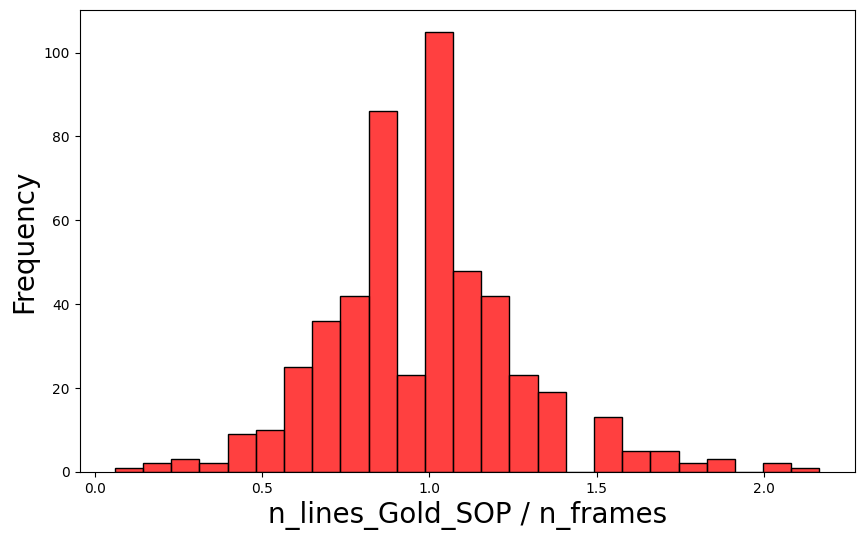}
        \caption{Histogram of ratios between number of lines of ground truth SOPs and the number of frames for each testing case.}
        \label{fig:ratio}
\end{figure}

\begin{figure}[H]
\centering
\includegraphics[width=0.7\textwidth]{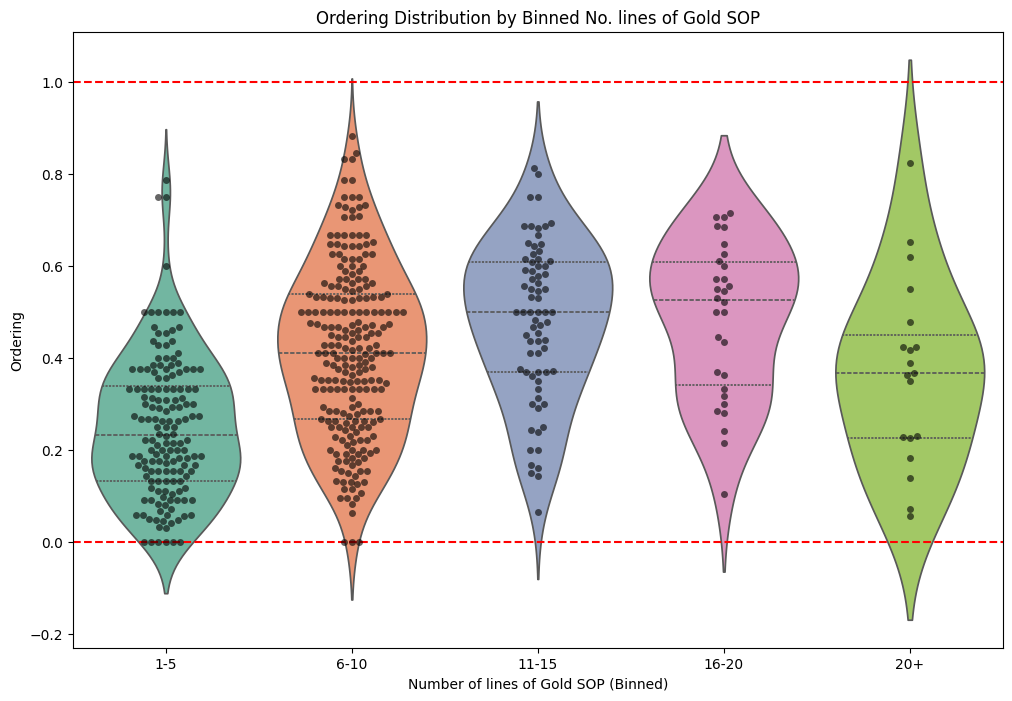}
\caption{Violins plots of binned temporal ordering according to gold SOP lengths.}
\label{fig:ordering_gt_sop_violin}
\end{figure}

 In Fig.\ref{fig:lines_sops}, we plotted the kernel density estimates of both the gold and predicted SOPs. The predicted SOPs have an average of 18.01 steps, with most concentrated around this value. In contrast, the gold SOPs average 8.44 steps, with a more even distribution across cases. This suggests that the model fails to capture the diversity seen in the gold SOPs, implying that it may have developed its own writing style to write the SOPs, that is different from the styles shown in the training examples. Additionally, we plotted the ratios between the number of SOP lines and the number of frames in Fig.\ref{fig:ratio}, observing a wide distribution. This raises the possibility that even among the ground truth SOPs, there may be inconsistencies in writing styles, hindering the effectiveness of in-context learning. We then further grouped the predictions according to the length of their ground truth SOPs in the following Fig.~\ref{fig:precision_gt_sop_violin}, \ref{fig:recall_gt_sop_violin}, \ref{fig:ordering_gt_sop_violin}. It is clear to see that, the more different between the lengths of the ground truth SOPs and the mean length of the predicted SOPs, the worse the performances. 

\begin{figure}[H]
\centering
\includegraphics[width=0.7\textwidth]{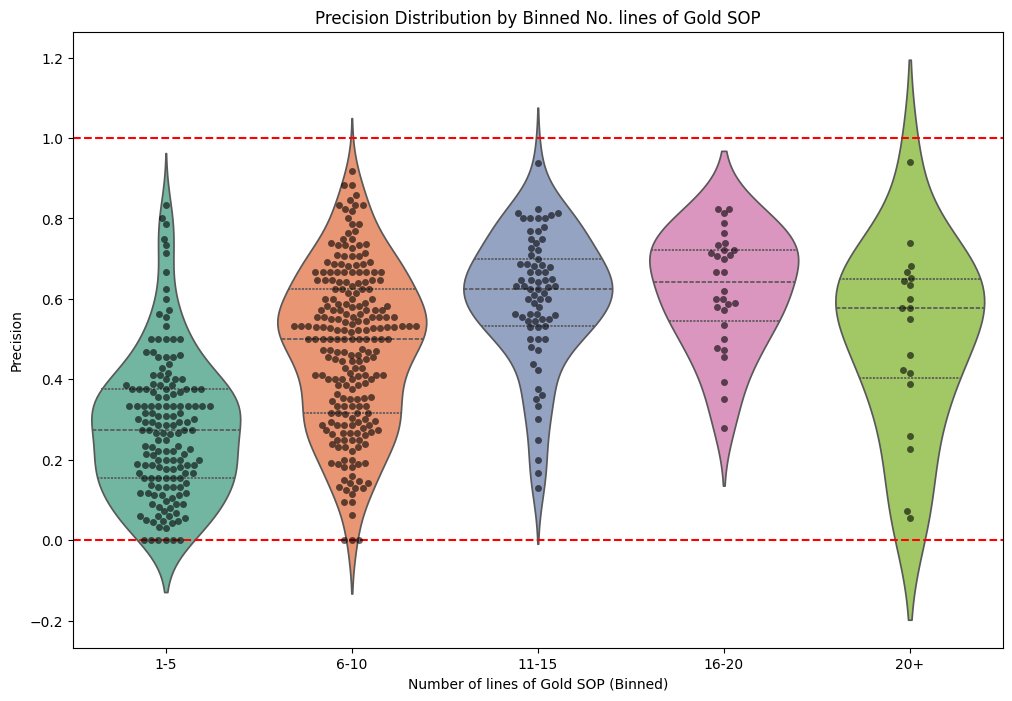}
\caption{Violins plots of binned precision according to gold SOP lengths.}
\label{fig:precision_gt_sop_violin}
\end{figure}

\begin{figure}[H]
\centering
\includegraphics[width=0.7\textwidth]{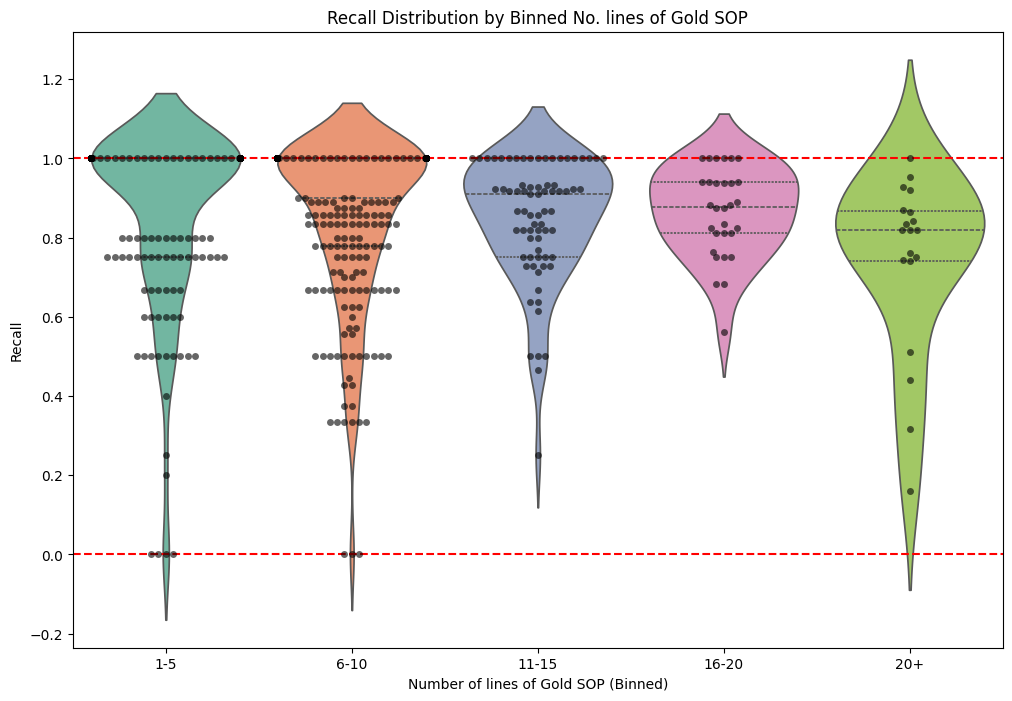}
\caption{Violins plots of binned recall according to gold SOP lengths.}
\label{fig:recall_gt_sop_violin}
\end{figure}

Based on our above observations, we propose following future research directions:
\begin{enumerate}
    \item Investigation into standardised writing style of SOPs is necessary. For example, one step of the SOP for each frame, with a fixed number of frames per video.
    \item Examination into the impact of the format of SOPs during prompt engineering might be beneficial.
    \item The current GPT-4o-mini based evaluation of temporal ordering is limited; a better metric or evaluation pipeline is required to further our insight into the impact of the models.
    \item Integration of other prompting techniques can further enhance the pipeline, such as self-reflection \cite{NEURIPS2023_1b44b878,NEURIPS2023_91edff07}.
\end{enumerate}

\section{Conclusion}

We conducted an exploratory evaluation of in-context learning for SOP generation and proposed an In-Context Ensemble (ICE) learning approach using pseudo-labels for SOP generation. Our preliminary results suggest that the proposed ICE pipeline is effective in enhancing foundational video-language models in SOP generation. However, we also observe that the current performance of available foundational models has not yet reached a satisfactory level in SOP generation, highlighting the challenges of low-level video understanding from visual inputs alone. 



\printbibliography

@misc{ye2024differential,
    title={Differential Transformer},
    author={Tianzhu Ye and Li Dong and Yuqing Xia and Yutao Sun and Yi Zhu and Gao Huang and Furu Wei},
    year={2024},
    eprint={2410.05258},
    archivePrefix={arXiv},
    primaryClass={cs.CL}
}

@article{DBLP:journals/corr/abs-1712-01815,
  author       = {David Silver and
                  Thomas Hubert and
                  Julian Schrittwieser and
                  Ioannis Antonoglou and
                  Matthew Lai and
                  Arthur Guez and
                  Marc Lanctot and
                  Laurent Sifre and
                  Dharshan Kumaran and
                  Thore Graepel and
                  Timothy P. Lillicrap and
                  Karen Simonyan and
                  Demis Hassabis},
  title        = {Mastering Chess and Shogi by Self-Play with a General Reinforcement
                  Learning Algorithm},
  journal      = {CoRR},
  volume       = {abs/1712.01815},
  year         = {2017},
  url          = {http://arxiv.org/abs/1712.01815},
  eprinttype    = {arXiv},
  eprint       = {1712.01815},
  bibsource    = {dblp computer science bibliography, https://dblp.org}
}

@misc{feng2023alphazerolike,
    title={Alphazero-like Tree-Search can Guide Large Language Model Decoding and Training},
    author={Xidong Feng and Ziyu Wan and Muning Wen and Stephen Marcus McAleer and Ying Wen and Weinan Zhang and Jun Wang},
    year={2023},
    eprint={2309.17179}
}

@misc{chen2024alphamath,
    title={AlphaMath Almost Zero: Process Supervision without Process},
    author={Guoxin Chen and Minpeng Liao and Chengxi Li and Kai Fan},
    year={2024},
    eprint={2405.03553},
    archivePrefix={arXiv},
    primaryClass={cs.CL}
}

@article{DBLP:journals/corr/abs-2110-14168,
  author       = {Karl Cobbe and
                  Vineet Kosaraju and
                  Mohammad Bavarian and
                  Mark Chen and
                  Heewoo Jun and
                  Lukasz Kaiser and
                  Matthias Plappert and
                  Jerry Tworek and
                  Jacob Hilton and
                  Reiichiro Nakano and
                  Christopher Hesse and
                  John Schulman},
  title        = {Training Verifiers to Solve Math Word Problems},
  journal      = {CoRR},
  volume       = {abs/2110.14168},
  year         = {2021},
  url          = {https://arxiv.org/abs/2110.14168},
  eprinttype    = {arXiv},
  eprint       = {2110.14168},
  bibsource    = {dblp computer science bibliography, https://dblp.org}
}

@inproceedings{wang-etal-2024-math,
    title = "Math-Shepherd: Verify and Reinforce {LLM}s Step-by-step without Human Annotations",
    author = "Wang, Peiyi  and
      Li, Lei  and
      Shao, Zhihong  and
      Xu, Runxin  and
      Dai, Damai  and
      Li, Yifei  and
      Chen, Deli  and
      Wu, Yu  and
      Sui, Zhifang",
    booktitle = "Proceedings of the 62nd Annual Meeting of the Association for Computational Linguistics (Volume 1: Long Papers)",
    year = "2024",
    publisher = "Association for Computational Linguistics",
    url = "https://aclanthology.org/2024.acl-long.510",
    doi = "10.18653/v1/2024.acl-long.510",
}

@inproceedings{lightman2024lets,
title={Let's Verify Step by Step},
author={Hunter Lightman and Vineet Kosaraju and Yuri Burda and Harrison Edwards and Bowen Baker and Teddy Lee and Jan Leike and John Schulman and Ilya Sutskever and Karl Cobbe},
booktitle={The Twelfth International Conference on Learning Representations},
year={2024},
url={https://openreview.net/forum?id=v8L0pN6EOi}
}

@inproceedings{asano2020self,
  title={Self-labelling via simultaneous clustering and representation learning},
  author={Asano, Yuki M. and Rupprecht, Christian and Vedaldi, Andrea},  
  booktitle={International Conference on Learning Representations (ICLR)},
  year={2020},
}

@InProceedings{Chen_2021_CVPR,
    author    = {Chen, Xinlei and He, Kaiming},
    title     = {Exploring Simple Siamese Representation Learning},
    booktitle = {Proceedings of the IEEE/CVF Conference on Computer Vision and Pattern Recognition (CVPR)},
    year      = {2021},
    pages     = {15750-15758}
}

@misc{gu2024mambalineartimesequencemodeling,
      title={Mamba: Linear-Time Sequence Modeling with Selective State Spaces}, 
      author={Albert Gu and Tri Dao},
      year={2024},
      eprint={2312.00752},
      archivePrefix={arXiv},
      primaryClass={cs.LG},
      url={https://arxiv.org/abs/2312.00752}, 
}

@inproceedings{NEURIPS2020_6b493230,
 author = {Lewis, Patrick and Perez, Ethan and Piktus, Aleksandra and Petroni, Fabio and Karpukhin, Vladimir and Goyal, Naman and K\"{u}ttler, Heinrich and Lewis, Mike and Yih, Wen-tau and Rockt\"{a}schel, Tim and Riedel, Sebastian and Kiela, Douwe},
 booktitle = {Advances in Neural Information Processing Systems},
 editor = {H. Larochelle and M. Ranzato and R. Hadsell and M.F. Balcan and H. Lin},
 pages = {9459--9474},
 publisher = {Curran Associates, Inc.},
 title = {Retrieval-Augmented Generation for Knowledge-Intensive NLP Tasks},
 url = {https://proceedings.neurips.cc/paper_files/paper/2020/file/6b493230205f780e1bc26945df7481e5-Paper.pdf},
 volume = {33},
 year = {2020}
}

@misc{kuratov2024search,
    title={In Search of Needles in a 11M Haystack: Recurrent Memory Finds What LLMs Miss},
    author={Yuri Kuratov and Aydar Bulatov and Petr Anokhin and Dmitry Sorokin and Artyom Sorokin and Mikhail Burtsev},
    year={2024},
    eprint={2402.10790},
    archivePrefix={arXiv},
    primaryClass={cs.CL}
}

@InProceedings{khan24a,
  title = 	 {Debating with More Persuasive {LLM}s Leads to More Truthful Answers},
  author =       {Khan, Akbir and Hughes, John and Valentine, Dan and Ruis, Laura and Sachan, Kshitij and Radhakrishnan, Ansh and Grefenstette, Edward and Bowman, Samuel R. and Rockt\"{a}schel, Tim and Perez, Ethan},
  booktitle = 	 {Proceedings of the 41st International Conference on Machine Learning},
  pages = 	 {23662--23733},
  year = 	 {2024},
  volume = 	 {235},
  series = 	 {Proceedings of Machine Learning Research},
  publisher =    {PMLR}
}

@misc{snell2024scaling,
    title={Scaling LLM Test-Time Compute Optimally can be More Effective than Scaling Model Parameters},
    author={Charlie Snell and Jaehoon Lee and Kelvin Xu and Aviral Kumar},
    year={2024},
    eprint={2408.03314},
    archivePrefix={arXiv},
    primaryClass={cs.LG}
}

@misc{saunders2022selfcritiquing,
    title={Self-critiquing models for assisting human evaluators},
    author={William Saunders and Catherine Yeh and Jeff Wu and Steven Bills and Long Ouyang and Jonathan Ward and Jan Leike},
    year={2022},
    eprint={2206.05802},
    archivePrefix={arXiv},
    primaryClass={cs.CL}
}

@inproceedings{yao2023tree,
title={Tree of Thoughts: Deliberate Problem Solving with Large Language Models},
author={Shunyu Yao and Dian Yu and Jeffrey Zhao and Izhak Shafran and Thomas L. Griffiths and Yuan Cao and Karthik R Narasimhan},
booktitle={Thirty-seventh Conference on Neural Information Processing Systems},
year={2023},
url={https://openreview.net/forum?id=5Xc1ecxO1h}
}

@inproceedings{wei2022chain,
title={Chain of Thought Prompting Elicits Reasoning in Large Language Models},
author={Jason Wei and Xuezhi Wang and Dale Schuurmans and Maarten Bosma and brian ichter and Fei Xia and Ed H. Chi and Quoc V Le and Denny Zhou},
booktitle={Advances in Neural Information Processing Systems},
editor={Alice H. Oh and Alekh Agarwal and Danielle Belgrave and Kyunghyun Cho},
year={2022},
url={https://openreview.net/forum?id=_VjQlMeSB_J}
}

@inproceedings{wang2023selfconsistency,
title={Self-Consistency Improves Chain of Thought Reasoning in Language Models},
author={Xuezhi Wang and Jason Wei and Dale Schuurmans and Quoc V Le and Ed H. Chi and Sharan Narang and Aakanksha Chowdhery and Denny Zhou},
booktitle={The Eleventh International Conference on Learning Representations },
year={2023},
url={https://openreview.net/forum?id=1PL1NIMMrw}
}

@inproceedings{NEURIPS2023_91edff07,
 author = {Madaan, Aman and Tandon, Niket and Gupta, Prakhar and Hallinan, Skyler and Gao, Luyu and Wiegreffe, Sarah and Alon, Uri and Dziri, Nouha and Prabhumoye, Shrimai and Yang, Yiming and Gupta, Shashank and Majumder, Bodhisattwa Prasad and Hermann, Katherine and Welleck, Sean and Yazdanbakhsh, Amir and Clark, Peter},
 booktitle = {Advances in Neural Information Processing Systems},
 editor = {A. Oh and T. Naumann and A. Globerson and K. Saenko and M. Hardt and S. Levine},
 pages = {46534--46594},
 publisher = {Curran Associates, Inc.},
 title = {Self-Refine: Iterative Refinement with Self-Feedback},
 url = {https://proceedings.neurips.cc/paper_files/paper/2023/file/91edff07232fb1b55a505a9e9f6c0ff3-Paper-Conference.pdf},
 volume = {36},
 year = {2023}
}

@InProceedings{pmlr-v202-von-oswald23a,
  title = 	 {Transformers Learn In-Context by Gradient Descent},
  author =       {Von Oswald, Johannes and Niklasson, Eyvind and Randazzo, Ettore and Sacramento, Joao and Mordvintsev, Alexander and Zhmoginov, Andrey and Vladymyrov, Max},
  booktitle = 	 {Proceedings of the 40th International Conference on Machine Learning},
  pages = 	 {35151--35174},
  year = 	 {2023},
  editor = 	 {Krause, Andreas and Brunskill, Emma and Cho, Kyunghyun and Engelhardt, Barbara and Sabato, Sivan and Scarlett, Jonathan},
  volume = 	 {202},
  series = 	 {Proceedings of Machine Learning Research},
  publisher =    {PMLR},
  url = 	 {https://proceedings.mlr.press/v202/von-oswald23a.html}
}

@inproceedings{pan-etal-2023-context,
    title = "What In-Context Learning {``}Learns{''} In-Context: Disentangling Task Recognition and Task Learning",
    author = "Pan, Jane  and
      Gao, Tianyu  and
      Chen, Howard  and
      Chen, Danqi",
    editor = "Rogers, Anna  and
      Boyd-Graber, Jordan  and
      Okazaki, Naoaki",
    booktitle = "Findings of the Association for Computational Linguistics: ACL 2023",
    month = jul,
    year = "2023",
    address = "Toronto, Canada",
    publisher = "Association for Computational Linguistics",
    url = "https://aclanthology.org/2023.findings-acl.527",
    doi = "10.18653/v1/2023.findings-acl.527",
    pages = "8298--8319"
}

@inproceedings{akyrek2023what,
title={What learning algorithm is in-context learning? Investigations with linear models},
author={Ekin Aky{\"u}rek and Dale Schuurmans and Jacob Andreas and Tengyu Ma and Denny Zhou},
booktitle={The Eleventh International Conference on Learning Representations },
year={2023},
url={https://openreview.net/forum?id=0g0X4H8yN4I}
}

@inproceedings{NEURIPS2023_0561738a,
 author = {Bietti, Alberto and Cabannes, Vivien and Bouchacourt, Diane and Jegou, Herve and Bottou, Leon},
 booktitle = {Advances in Neural Information Processing Systems},
 editor = {A. Oh and T. Naumann and A. Globerson and K. Saenko and M. Hardt and S. Levine},
 pages = {1560--1588},
 publisher = {Curran Associates, Inc.},
 title = {Birth of a Transformer: A Memory Viewpoint},
 url = {https://proceedings.neurips.cc/paper_files/paper/2023/file/0561738a239a995c8cd2ef0e50cfa4fd-Paper-Conference.pdf},
 volume = {36},
 year = {2023}
}

@inproceedings{wei2022finetuned,
title={Finetuned Language Models are Zero-Shot Learners},
author={Jason Wei and Maarten Bosma and Vincent Zhao and Kelvin Guu and Adams Wei Yu and Brian Lester and Nan Du and Andrew M. Dai and Quoc V Le},
booktitle={International Conference on Learning Representations},
year={2022},
url={https://openreview.net/forum?id=gEZrGCozdqR}
}

@InProceedings{10.1007/978-3-031-16443-9_56,
author="Xu, Mou-Cheng
and Zhou, Yukun
and Jin, Chen
and de Groot, Marius
and Alexander, Daniel C.
and Oxtoby, Neil P.
and Hu, Yipeng
and Jacob, Joseph",
editor="Wang, Linwei
and Dou, Qi
and Fletcher, P. Thomas
and Speidel, Stefanie
and Li, Shuo",
title="Bayesian Pseudo Labels: Expectation Maximization for Robust and Efficient Semi-supervised Segmentation",
booktitle="Medical Image Computing and Computer Assisted Intervention -- MICCAI 2022",
year="2022",
publisher="Springer Nature Switzerland",
address="Cham",
pages="580--590"
}

@article{XU2024103125,
title = {Expectation maximisation pseudo labels},
journal = {Medical Image Analysis},
volume = {94},
pages = {103125},
year = {2024},
issn = {1361-8415},
doi = {https://doi.org/10.1016/j.media.2024.103125},
url = {https://www.sciencedirect.com/science/article/pii/S1361841524000501},
author = {Moucheng Xu and Yukun Zhou and Chen Jin and Marius {de Groot} and Daniel C. Alexander and Neil P. Oxtoby and Yipeng Hu and Joseph Jacob}
}

@inproceedings{NEURIPS2023_1b44b878,
 author = {Shinn, Noah and Cassano, Federico and Gopinath, Ashwin and Narasimhan, Karthik and Yao, Shunyu},
 booktitle = {Advances in Neural Information Processing Systems},
 editor = {A. Oh and T. Naumann and A. Globerson and K. Saenko and M. Hardt and S. Levine},
 pages = {8634--8652},
 publisher = {Curran Associates, Inc.},
 title = {Reflexion: language agents with verbal reinforcement learning},
 url = {https://proceedings.neurips.cc/paper_files/paper/2023/file/1b44b878bb782e6954cd888628510e90-Paper-Conference.pdf},
 volume = {36},
 year = {2023}
}

@inproceedings{Lee2013PseudoLabelT,
  title={Pseudo-Label : The Simple and Efficient Semi-Supervised Learning Method for Deep Neural Networks},
  author={Dong-Hyun Lee},
  year={2013},
  url={https://api.semanticscholar.org/CorpusID:18507866}
}

@inproceedings{NEURIPS2020_1457c0d6,
 author = {Brown, Tom and Mann, Benjamin and Ryder, Nick and Subbiah, Melanie and Kaplan, Jared D and Dhariwal, Prafulla and Neelakantan, Arvind and Shyam, Pranav and Sastry, Girish and Askell, Amanda and Agarwal, Sandhini and Herbert-Voss, Ariel and Krueger, Gretchen and Henighan, Tom and Child, Rewon and Ramesh, Aditya and Ziegler, Daniel and Wu, Jeffrey and Winter, Clemens and Hesse, Chris and Chen, Mark and Sigler, Eric and Litwin, Mateusz and Gray, Scott and Chess, Benjamin and Clark, Jack and Berner, Christopher and McCandlish, Sam and Radford, Alec and Sutskever, Ilya and Amodei, Dario},
 booktitle = {Advances in Neural Information Processing Systems},
 editor = {H. Larochelle and M. Ranzato and R. Hadsell and M.F. Balcan and H. Lin},
 pages = {1877--1901},
 publisher = {Curran Associates, Inc.},
 title = {Language Models are Few-Shot Learners},
 url = {https://proceedings.neurips.cc/paper_files/paper/2020/file/1457c0d6bfcb4967418bfb8ac142f64a-Paper.pdf},
 volume = {33},
 year = {2020}
}

@InProceedings{Chen_2024_CVPR,
    author    = {Chen, Zhe and Wu, Jiannan and Wang, Wenhai and Su, Weijie and Chen, Guo and Xing, Sen and Zhong, Muyan and Zhang, Qinglong and Zhu, Xizhou and Lu, Lewei and Li, Bin and Luo, Ping and Lu, Tong and Qiao, Yu and Dai, Jifeng},
    title     = {InternVL: Scaling up Vision Foundation Models and Aligning for Generic Visual-Linguistic Tasks},
    booktitle = {Proceedings of the IEEE/CVF Conference on Computer Vision and Pattern Recognition (CVPR)},
    year      = {2024},
    pages     = {24185-24198}
}

@InProceedings{Lin_2024_CVPR,
    author    = {Lin, Ji and Yin, Hongxu and Ping, Wei and Molchanov, Pavlo and Shoeybi, Mohammad and Han, Song},
    title     = {VILA: On Pre-training for Visual Language Models},
    booktitle = {Proceedings of the IEEE/CVF Conference on Computer Vision and Pattern Recognition (CVPR)},
    year      = {2024},
    pages     = {26689-26699}
}

@inproceedings{maaz-etal-2024-video,
    title = "Video-{C}hat{GPT}: Towards Detailed Video Understanding via Large Vision and Language Models",
    author = "Maaz, Muhammad  and
      Rasheed, Hanoona  and
      Khan, Salman  and
      Khan, Fahad",
    editor = "Ku, Lun-Wei  and
      Martins, Andre  and
      Srikumar, Vivek",
    booktitle = "Proceedings of the 62nd Annual Meeting of the Association for Computational Linguistics (Volume 1: Long Papers)",
    month = aug,
    year = "2024",
    address = "Bangkok, Thailand",
    publisher = "Association for Computational Linguistics",
    url = "https://aclanthology.org/2024.acl-long.679",
    pages = "12585--12602"
}

@inproceedings{
roy2024consistencyguided,
title={Consistency-guided Prompt Learning for Vision-Language Models},
author={Shuvendu Roy and Ali Etemad},
booktitle={The Twelfth International Conference on Learning Representations},
year={2024},
url={https://openreview.net/forum?id=wsRXwlwx4w}
}

@misc{wornow2024multimodal,
    title={Do Multimodal Foundation Models Understand Enterprise Workflows? A Benchmark for Business Process Management Tasks},
    author={Michael Wornow and Avanika Narayan and Ben Viggiano and Ishan S. Khare and Tathagat Verma and Tibor Thompson and Miguel Angel Fuentes Hernandez and Sudharsan Sundar and Chloe Trujillo and Krrish Chawla and Rongfei Lu and Justin Shen and Divya Nagaraj and Joshua Martinez and Vardhan Agrawal and Althea Hudson and Nigam H. Shah and Christopher Re},
    year={2024},
    eprint={2406.13264},
    archivePrefix={arXiv},
    primaryClass={cs.AI}
}

@article{wei2022emergent,
  title={Emergent abilities of large language models},
  author={Wei, Jason and Tay, Yi and Bommasani, Rishi and Raffel, Colin and Zoph, Barret and Borgeaud, Sebastian and Yogatama, Dani and Bosma, Maarten and Zhou, Denny and Metzler, Donald and others},
  journal={arXiv preprint arXiv:2206.07682},
  year={2022}
}

@article{midl2022_xu,
  author    = {Mou-Cheng Xu and
               Yukun Zhou and
               Chen Jin and
               Stefano B. Blumberg and
               Frederick J. Wilson and
               Marius de Groot and
               Daniel C. Alexander and
               Neil P. Oxtoby and
               Joseph Jacob
               },
  title     = {Learning Morphological Feature Perturbations for Calibrated Semi-Supervised Segmentation},
  journal   = {International Conference on Medical Imaging with Deep Learning (MIDL)},
  year      = {2022},
}

@article{tmi2023_xu,
  author    = {Mou-Cheng Xu and
               Yukun Zhou and
               Chen Jin and
               Marius de Groot and
               Daniel C. Alexander and
               Neil P. Oxtoby and
               Joseph Jacob
               },
  title     = {MisMatch: Calibrated Segmentation via Consistency on Differential Morphological Feature Perturbations With Limited Labels},
  journal   = {IEEE Transactions on Medical Imaging (TMI)},
  year      = {2023},
}

@article{fixmatch2020,
  author    = {Kihyuk Sohn and
               David Berthelot and
               Chun-Liang Li and
               Zizhao Zhang and
               Nicholas Carlini and
               Ekin D. Cubuk and
               Alex Kurakin and
               Han Zhang and
               Colin Raffel},
  title     = {FixMatch: Simplifying Semi-Supervised Learning with Consistency and Confidence},
  journal   = {Neural Information Processing Systems (NeurIPS)},
  year      = {2020},
}
\clearpage

\appendix
\label{appendix}
\section{A successful case}
\begin{figure}[H]
    \centering
    \begin{minipage}[t][0.9\textheight][t]{\textwidth} 
    \footnotesize
        \begin{multicols}{2}
        \centering\textbf{Ground Truth}\\[1em]
        \raggedright
        \begin{enumerate}
            \item Click the purple/gray search bar on the top left of the screen (icon with three lines stacked).
            \item Type ‘millennials-to-snake-people’.
            \item Press enter.
            \item Click the first pane on the top of the field of panes of projects that show up in the middle, with black text that has words like timeline, item, management, extension.
            \item Click the black bold “Project Information” tab on the top left side of the screen above “Repository.”
            \item Click the black bold “Members” text on the left side of the screen under “Labels.”
            \item Click the blue “Invite members” button on the top right side of the screen.
            \item Type "yjlou."
            \item Click the pane with the colored avatar that pops up below.
            \item Click the right side of the text box that says “Username or email address.”
            \item Type "a11yproject."
            \item Click the pane with the colored avatar that pops up below.
            \item Click the right side of the text box that says “Username or email address.”
            \item Click the dropdown underneath the black text saying “Select a role.”
            \item Click “Reporter” from the dropdown.
            \item Click the blue “Invite” button on the bottom right.
        \end{enumerate}

        \columnbreak

        \centering\textbf{Prediction}\\[1em]
        \raggedright
        \begin{enumerate}
            \item Navigate to Projects: Click on the "Projects" tab located in the top navigation bar of the GitLab dashboard, positioned next to the "Dashboard" tab.
            \item Search for Project: In the search bar at the top of the Projects page, which is labeled "Search GitLab," click to focus on it.
            \item Type Project Name: Type "millennials-to-snake-people" into the search bar.
            \item Select Project: Wait for the suggestions to load and then click on the project name "millennials-to-snake-people" from the dropdown suggestions to navigate to the project's main page.
            \item Access Members Section: In the left sidebar of the project page, click on "Project information" to expand the section.
            \item Click Members: Under the expanded "Project information" section, click on the "Members" option to go to the members page.
            \item Open Invite Members Dialog: In the Members section, locate and click the blue "Invite members" button on the top right side of the screen to open the invite members dialog.
            \item Focus on Username Field: In the "Invite members" pop-up window, click into the text box labeled "Username or email address."
            \item Type Username: Type "yjlou" into the text box to search for the user you want to invite.
            \item Select User from Results: Wait for the system to display the search results. When "yjlou" appears in the results, click on it to select that user.
            \item Open Role Dropdown: Below the selected username, click on the dropdown menu labeled "Select a role" to choose the role for the invited member.
            \item Choose Role: From the dropdown options, select "Reporter" by clicking on it to set the user's access level.
            \item Leave Expiration Date Blank: If necessary, locate the "Access expiration date (optional)" field and leave it blank.
            \item Send Invitation: Click the blue "Invite" button located at the bottom right of the pop-up window to send the invitation.
            \item Confirm Invitation Success: After the invitation is sent, observe the confirmation message that appears, indicating that the member has been successfully added.
            \item Review Members List: Finally, review the list of project members to ensure that "yjlou" now appears in the members' list, confirming the successful invitation.
        \end{enumerate}
        \end{multicols}
    \end{minipage}
    \caption{Comparison of the Ground Truth with the Prediction of "578-2024-01-07-17-23-54". Precision: 81.25\%. Recall: 93.75\%. Temporal order: 68.75\%.}
    \label{fig:comparison_good}
\end{figure}

\begin{figure}[H]
\centering
\includegraphics[width=0.8\textwidth]{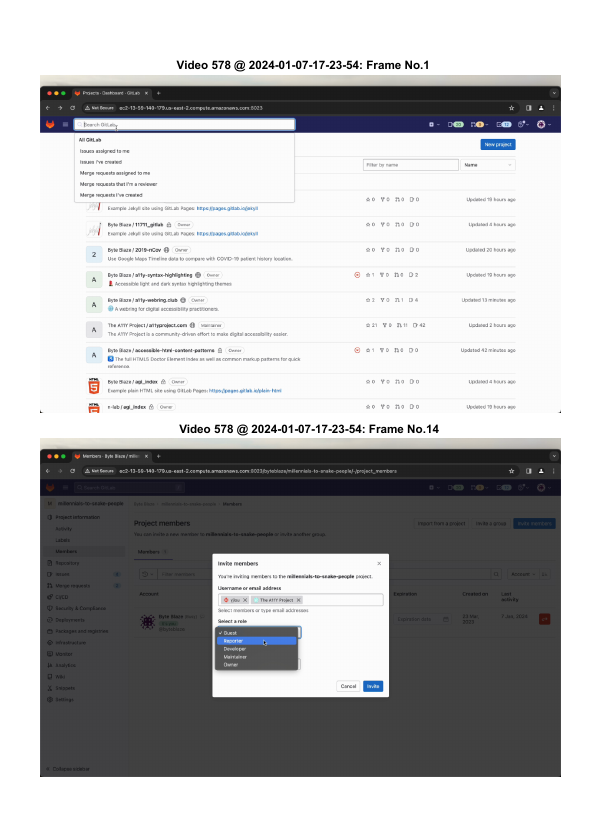}
\caption{Testing video (578-2024-01-07-17-23-54). Row 1: 1st frame from the video. Row 2: 9th frame from the video.}
\label{fig:screenshots1}
\end{figure}

The task "578-2024-01-07-17-23-54" in Fig. \ref{fig:screenshots1} involves adding new users to a GitLab repository as reporters. The model used was ICE with GPT-4o-mini. As shown in Fig. \ref{fig:comparison_good}, all steps of the ground truth were correctly predicted, but the prediction included extra details, highlighted in red. The metrics calculated by GPT-4o-mini for this task are: precision 81.25\%, recall 93.75\%, and temporal order correctness 68.75\%. We observed that both workflows actually follow the same temporal order, though they are not perfectly synchronised. This highlights a limitation in the use of GPT-4o-mini for evaluation, suggesting that the model's foundational ability may lack some basic temporal understanding. Consequently, the model's performance may not be as poor as the metrics indicate.

\section{A failed case}
\begin{figure}[H]
    \centering
    \begin{minipage}[t][0.8\textheight][t]{0.9\textwidth} 
        \begin{multicols}{2}
        \centering\textbf{Ground Truth}\\[1em]
        \raggedright
        \begin{enumerate}
            \item Click on "Explore" under the "Projects" heading
            \item Click on the search bar in the top right
            \item Type in "metaseq"
            \item From the results below, click on "Administrator / metaseq"
            \item Click on the "Repository" tab in the left sidebar.
            \item Click on the "Contributors" tab in the left sidebar.
            \item Report the information of the top three people listed (top left, top right, second left)
        \end{enumerate}

        \columnbreak

        \centering\textbf{Prediction}\\[1em]
        \raggedright
        \begin{enumerate}
            \item On the GitLab dashboard, locate the "Projects" tab in the top navigation bar, situated between the "Dashboard" and "Groups" options, and click on it.
            \item In the "Projects" section, observe the list of projects displayed. Scroll down until you can see the project titled "The A11Y Project / a11yproject.com", which is the fourth project from the top.
            \item Click on the project titled "The A11Y Project / a11yproject.com" to navigate to its main page.
            \item Once on the project page, locate the "Project Information" tab on the top left side, directly under the project name, and click on it.
            \item In the expanded options under "Project Information," click on the "Members" tab, located directly beneath the "Project Information" heading.
            \item Click the blue “Invite members” button located at the top right of the members section, next to the “Invite a group” option.
            \item In the popup window that appears, click on the text box labeled "Username or email address."
            \item Type "Koushik" into the "Username or email address" text box.
            \item Press Enter to select "Koushik Dutra" from the dropdown that appears as you type.
            \item Click on the dropdown labeled "Select a role" located directly beneath the username text box.
            \item Choose "Guest" from the options available in the dropdown list.
            \item Click the blue "Invite" button located at the bottom right of the popup window inside the shaded area to send the invitation.
            \item Confirm that the member has been successfully added by checking for a notification on the screen that states "Members were successfully added."
        \end{enumerate}
        \end{multicols}
    \end{minipage}
    \caption{Comparison of the Ground Truth with the Prediction of "317-2024-02-17-14-32-37". Precision: 0\%. Recall: 0\%. Temporal order: 0\%.}
    \label{fig:comparison_bad}
\end{figure}

\begin{figure}[H]
\centering
\includegraphics[width=\textwidth]{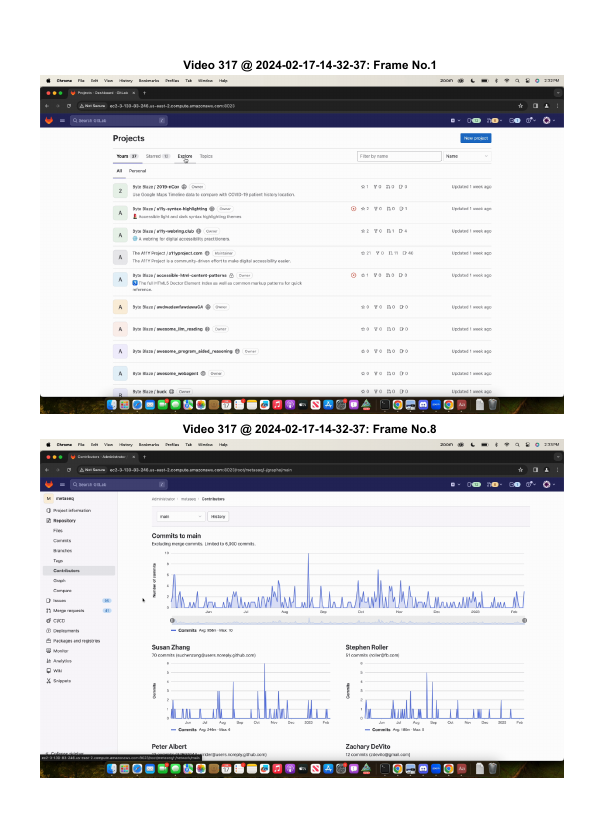}
\caption{Testing video (317-2024-02-17-14-32-37). Row 1: 1st frame from the video. Row 2: 8th last frame from the video.}
\label{fig:screenshots2}
\end{figure}

Fig. \ref{fig:screenshots2} contains screenshots of task "317-2024-02-17-14-32-37," which was tested with GPT-4o-mini; the model hallucinated the SOP. As shown in Fig. \ref{fig:comparison_bad}, this task involves listing the names and number of commits of the top three contributors to a GitLab repository called "metaseq." However, the model's prediction is a completely different SOP for adding new guests to the GitLab repository. This indicates that the model does not fundamentally understand the software.

\end{document}